\documentclass[letterpaper, 10 pt, conference]{ieeeconf}  % Comment this line out if you need a4paper

\IEEEoverridecommandlockouts                              % This command is only needed if 
\usepackage{graphics} % for pdf, bitmapped graphics files
\usepackage{epsfig} % for postscript graphics files
\usepackage{caption}
\usepackage{subcaption}
\usepackage{xcolor}
\usepackage{color}
\usepackage{tabularx} % in the preamble

\usepackage{booktabs}
\usepackage{siunitx}

\title{\LARGE \bf
Benchmarking Off-The-Shelf Solutions to Robotic Assembly Tasks
}

\author{Wenzhao Lian$^{1}$, Tim Kelch$^{1}$, Dirk Holz$^{1}$,  Adam Norton$^{2}$, and Stefan Schaal$^{1}$% <-this % stops a space
\thanks{$^{1}$X, The Moonshot Factory, Mountain View, CA 94043 USA. {\tt\small \{wenzhaol, tkelch, holz, sschaal\}@x.team}}%
\thanks{$^{2}$New England Robotics Validation and Experimentation (NERVE) Center,
University of Massachusetts Lowell,
Lowell, MA 01853 USA {\tt\small adam\_noron@uml.edu}}%
}

\begin{document}

\maketitle
\thispagestyle{empty}
\pagestyle{empty}

%%%%%%%%%%%%%%%%%%%%%%%%%%%%%%%%%%%%%%%%%%%%%%%%%%%%%%%%%%%%%%%%%%%%%%%%%%%%%%%%
\begin{abstract}

In recent years, many learning based approaches have been studied to realize robotic manipulation and assembly tasks, often including vision and force/tactile feedback. However, it remains frequently unclear what is the baseline state-of-the-art performance and what are the bottleneck problems. In this work, we evaluate some off-the-shelf (OTS) industrial solutions on a recently introduced benchmark, the National Institute of Standards and Technology (NIST) Assembly Task Boards. A set of assembly tasks are introduced and baseline methods are provided to understand their intrinsic difficulty. Multiple sensor-based robotic solutions are then evaluated, including hybrid force/motion control and 2D/3D pattern matching algorithms. An end-to-end integrated solution that accomplishes the tasks is also provided. The results and findings throughout the study reveal a few noticeable factors that impede the adoptions of the OTS solutions: expertise dependent, limited applicability, lack of interoperability, no scene awareness or error recovery mechanisms, and high cost. This paper also provides a first attempt of an objective benchmark performance on the NIST Assembly Task Boards as a reference comparison for future works on this problem.
% In recent years, many learning based approaches have been studied to realize robotic manipulation and assembly tasks, often including vision and force/tactile feedback. However, it remains frequently unclear what is the baseline state-of-the-art performance and what are the bottleneck problems. In this work, we evaluate some off-the-shelf (OTS) industrial solutions on a recently introduced benchmark, the National Institute of Standards and Technology (NIST) challenge on ``Robotics Grasping and Manipulation for Assembly" (Task Board #1). A set of assembly tasks are introduced and baseline methods are provided to understand their intrinsic difficulty. Multiple sensor-based robotic solutions are then evaluated, including hybrid force/motion control and 2D/3D pattern matching algorithms. An end-to-end integrated solution that accomplishes the tasks is also provided. The results and findings throughout the study reveal a few noticeable factors that impede the adoptions of the OTS solutions: expertise dependent, limited applicability, lack of interoperability, no scene awareness or error recovery mechanisms, and high cost. This paper also provides a first attempt of an objective benchmark performance on the NIST challenge as a reference comparison for other work on the challenge. 

\end{abstract}

\section{INTRODUCTION}

In recent years, robotic manipulation and assembly tasks have attracted significant attention from the robotics research community, in part motivated by the increasing interest of AI-enabled robotics for manufacturing. Various lines of work have been pursued, including end-to-end reinforcement learning~\cite{inoue2017deep}, learning from demonstration~\cite{wu2020learning}, structured parameter learning~\cite{Voigt2020learning}, and reusable primitive prototyping~\cite{nagele2018prototype}. Promising results have been demonstrated in peg-in-hole type of assembly tasks, e.g., with $\#x$ numbers of demonstrations and/or self-play trials, $y\%$ success rate is achieved. However, OTS solutions combined with heavy customization are still the go-to solution in manufacturing assembly tasks in industry; in contrast, in the research community, OTS solutions are rarely considered as baseline methods due to factors including cost, accessibility and interoperability. The research community often still lacks access to knowledge of where the OTS state-of-the-art stands and what key metrics are expected to be improved.

Benchmarks and challenges have been proven beneficial in many problems, such as ImageNet~\cite{deng2009imagenet} for computer vision or the Amazon Picking Challenge~\cite{correll2016analysis} for grasping. Even though various challenges have been proposed for assembly problems in the last few years, e.g., the World Robot Summit (WRS) Assembly Challenge~\cite{yokokohji2019assembly, von2020robots}, and the NIST Assembly Task Boards~\cite{falco2018performance}, there hasn't been a common agreement on which parts of the assembly tasks are unsolved or difficult. One reason is that most participants from the robotics or learning community are limited by low cost hardware (e.g., consumer-level cameras like Intel Realsense and Microsoft Kinect) and non performance focused tooling (e.g., low precision calibration packages). It is not obvious to the community what are limiting factors to the end-to-end performance and what is the state-of-the-art with mature OTS tools. Therefore, the goal of this work is to benchmark a number of OTS solutions and gain an understanding of their strengths and weaknesses in assembly tasks. Investigating the performance of OTS solutions not only reveals their limitations, but also suggests future research directions and relevant metrics.

A subset of assembly tasks in the NIST Assembly Task Boards were selected as the benchmarking testbed, focusing on peg-in-hole type of operations as insertion tasks account for the largest number of assembly operations~\cite{falco2018performance}. Specifically, 12 parts were chosen, ranging from round pegs to multi-pin connectors. We mostly adopted the experimental protocol suggested in~\cite{falco2018performance} and designed customized fixtures to allow experiments to be reproducible and quantified. For each of the 12 tasks, we implemented 3 solutions using OTS tools: 1) a position control baseline, 2) a hybrid motion/force control method (spiral search), and 3) a 3D vision-based pose estimation method. The benchmarking experiments are further extended by introducing extra workspace variations to characterize the robustness of each method, including variations in task board presentation and external lighting.

Based on the empirical evaluation, we found the OTS tools are capable to solve most of the tasks, with different levels of generalization. Meanwhile, there are a few noticeable factors that impede the OTS solutions from being widely adopted: 1) expertise dependent: parameter tuning is required from task to task, in both force and vision based methods; 2) limited applicability: e.g., vision methods based on 3D cameras fail at a number of scenarios such as transparent assembly parts; 3) lack of interoperability: e.g., if the native support of integrating a robot and a camera is not provided, obtaining an accurate hand-eye calibration is challenging; 4) no scene awareness or error recovery mechanism: e.g., not robust to anomalies such as working pieces obscured by or entangled with cables; 5) high cost.

\section{Related Work}

In this section, we briefly review some of the relevant research in robotic assembly and the recently proposed NIST Assembly Task Boards. We also give a summary of previous benchmarking efforts in the related field.

Assembly tasks have attracted increasing attention in the robotics and learning community. Particularly in the field of deep RL, \cite{inoue2017deep} proposed a deep RL approach that takes robot proprioception signals and force-torque feedback as input, to solve a peg-in-hole task. Similarly, \cite{vecerik2018leveraging} examined learning from demonstration with deep learning including camera-based inputs for insertion tasks. \cite{thomas2018learning} studied how to leverage prior information such as the workpieces' CAD models to guide the RL agent, to more efficiently tackle assembly tasks including peg-in-hole and gear assembly. \cite{wu2020learning} further proposed to learn the reward via demonstration and sampling, which is later used for RL training on peg-in-hole and USB connector insertion tasks. More structured learning approaches including \cite{Voigt2020learning} are studied leveraging dynamical systems and black-box function optimization, achieving promising results on a key-in-lock problem.

Even though multiple lines of research have been pursued to address the assembly tasks, there hasn't been a widely adopted benchmark nor clearly reproducible baselines. For example, ablation studies were conducted in \cite{thomas2018learning} and other learning-based methods to serve as baselines in \cite{wu2020learning, Voigt2020learning}. \cite{inoue2017deep} quoted the maximum error that can be handled by an OTS force-based solution from its product specifications, but it is not clear how the OTS solution actually works on particular tasks.

The NIST Assembly Task Boards have been used in multiple competitions such as the IROS Robotic Grasping and Manipulation Competition~\cite{falco2018performance}, and has been receiving growing attention~\cite{watson2020autonomous, gorjupcombining, carfi2020rss}. With a standardized and reproducible benchmark kit, it contains a rich set of operations, such as peg-in-hole, connector insertion, gear meshing, and nut threading. Existing techniques such as vision-based contour matching and force-based spiral searching have been attempted on a few operations including peg-in-hole tasks~\cite{watson2020autonomous, gorjupcombining}. However, standalone implementations of algorithms without comparison, combined with other module choices such as opensource calibration packages and consumer-level cameras, hinders the understanding of maturity of autonomous assembly solutions, in particular relative to established high quality OTS solutions.

There have been a few focused attempts benchmarking OTS solutions on object detection and localization~\cite{vision_benchmark2012}, and force-based robot control~\cite{force_benchmark2016}. \cite{vision_benchmark2012} evaluated three OTS machine vision tools for detecting and localizing objects in grayscale images. \cite{force_benchmark2016} did a detailed analysis of two OTS force control solutions in terms of settling stability and disturbance handling. A related functional level evaluation of autonomous assembly solutions is \cite{von2020robots}, where the solutions from six teams participating in the WRS 2018 Assembly Challenge are summarized in a high level. As the challenge covered a wide range of tasks and the teams had the flexibility in customizing all modules in mechanics, eletronics, and software, results are more valuable as a comparison at the task-level rather than component-level. Insights were provided by comparing the diversity of chosen approaches and their overall performance, including the low correlation between algorithm novelty and task success, and lack of error recovery mechanisms in general. 

In contrast to previous work, our investigation focuses on the same peg-in-hole insertion tasks but with diverse work pieces and different workspace setups, in order to compare different OTS solutions in terms of reliability and usability. We therefore provide a first attempt on an objective baseline performance on the NIST Assembly Task Boards for insertion tasks that could serve as a comparison for future works on this problem.

\section{Benchmark Design}
\label{section: benchmark}

In this section, we introduce the task setup, experimental protocol, and the metrics considered.

\subsection{Tasks}
\label{section: tasks}

A subset of insertion tasks are selected from the NIST Assembly Task Board \#1, including 4 round pegs (with diameters 4mm, 8mm, 12mm, and 16mm), 4 rectangle pegs (with side lengths 4x4mm, 8x7mm, 12x8mm, and 16x10mm), and 4 connectors (USB, RJ45, waterproof, and DSUB). These parts all require peg-in-hole type of insertion, ranging from easier high clearance insertion to harder multi-pin alignment. Other parts that require more complex operations such as twisting in gear assembly are deferred to future work. For naming simplicity, throughout the rest of the paper, we use \textit{pegs} and \textit{holes}, to refer to the pegs/connectors and holes/sockets.

\begin{figure}%[!htb]
	\centering
	\includegraphics[width=0.47\columnwidth]{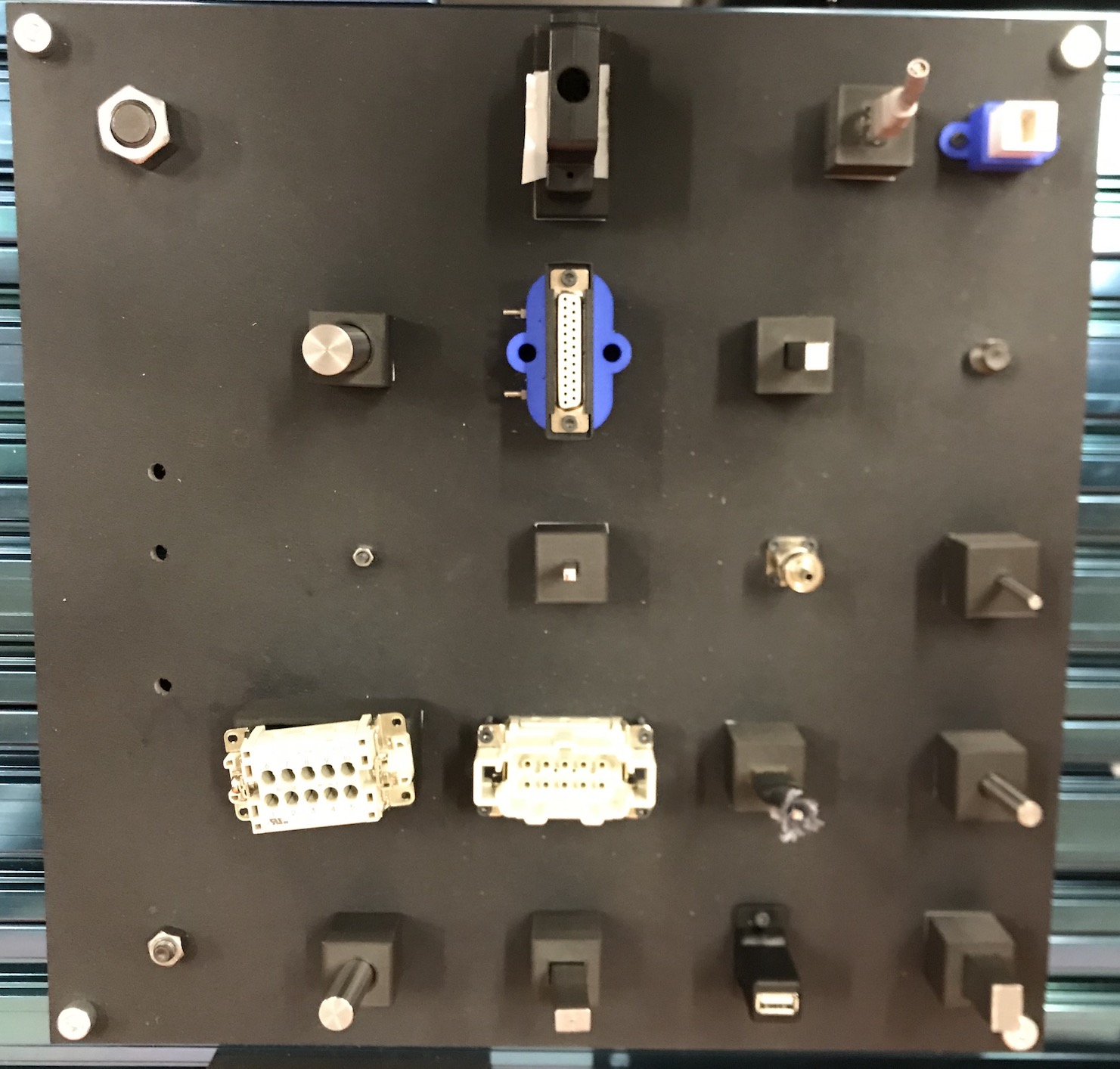}
	\hfill
	\includegraphics[width=0.47\columnwidth]{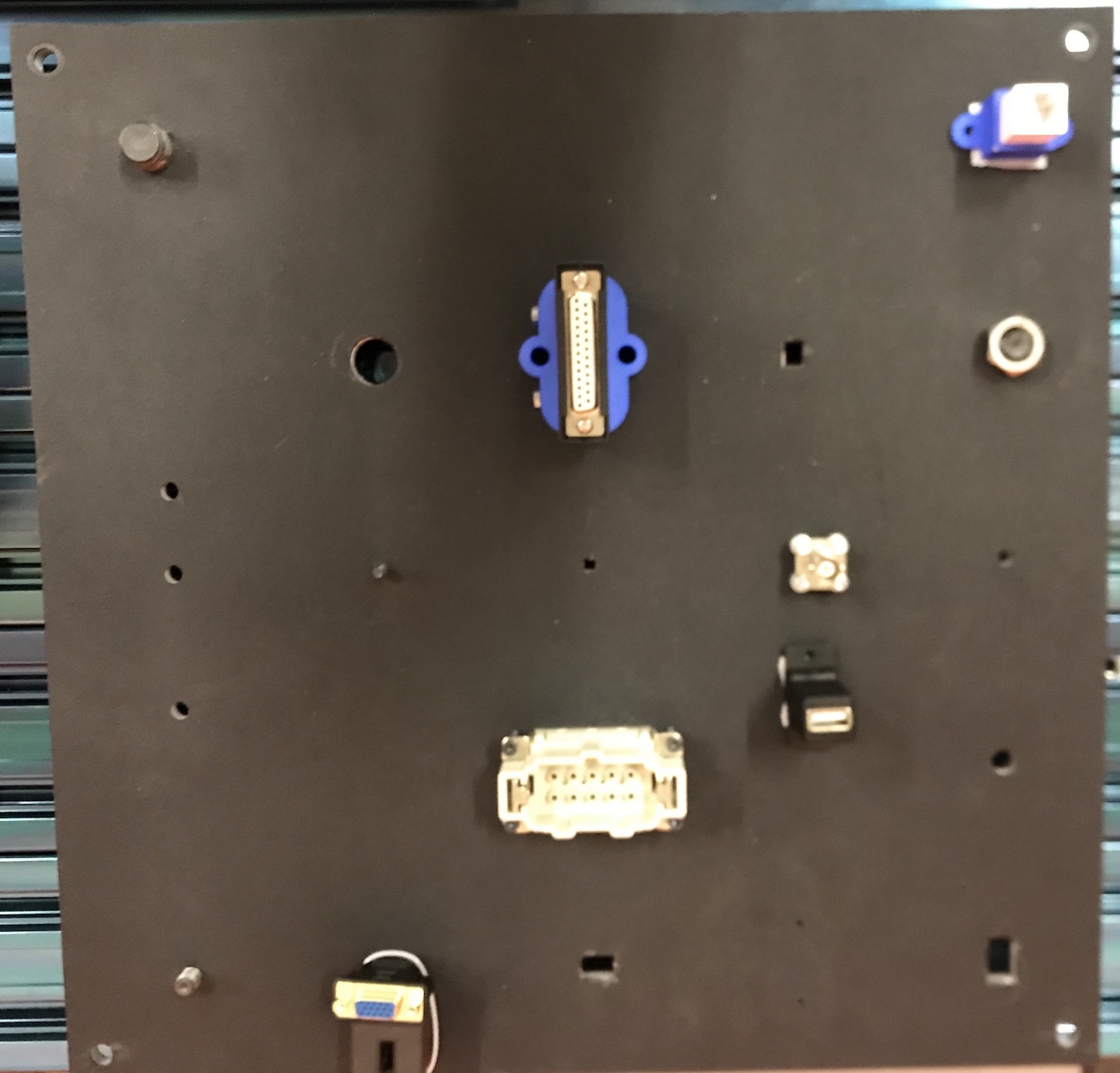}
	\caption{Illustration of the task setup. Left: nest board; right: assembly board.}
	\label{fig:experiment:task_setup}
	\vspace{-5mm}
\end{figure}

A typical task setup is illustrated in Fig.~\ref{fig:experiment:task_setup}, where two NIST boards are placed within the robot workspace, one to grasp pegs from (nest board) and the other to insert pegs into (assembly board). The boards are modified from the original plexiglass material to aluminum with a matte black plastic sheet on top, to reduce the wear-and-tear of the boards and mitigate any artificial reflection. In addition, the board corners are adjusted such that the boards can be mounted straight to the worktable extrusions. The assembly board is rigidly mounted or can be moved depending on the particular workspace variation as shown in Fig.~\ref{fig:experiment:task_variation}. Meanwhile, the nest board is always rigidly mounted on the worktable; the nests that hold pegs ensure pegs are facing downward, i.e., no regrasping is required for insertion after grasping. The nest dimensions are designed such that they allow slight deviations from the \textit{nominal} grasping pose: +/-3mm in translation and approximately +/-5degrees in orientation.

The end-to-end experimental procedure for a peg-hole pair is summarized as follows.
\begin{enumerate}
    \item Optionally, the operator introduces a disturbance of the setup such as turning on the external light (Fig.~\ref{fig:experiment:task_variation}c) or changing the assembly board resting configuration (Fig.~\ref{fig:experiment:task_variation}d).
    \item The operator places the peg within its nest, allowing a small deviation from the \textit{nominal} pose in both translation and orientation.
    % \adam{What characteristic is randomized? Orientation? May want this to be more specific. Although the previous paragraph said they are always facing downward, so is any part of it actually randomized?}
    \item The robot moves to a predefined grasping configuration (corresponding to the nominal peg resting pose) to grasp the peg.
    \item Optionally, the peg-in-gripper and target hole poses are estimated by a vision module.
    \item The robot attempts to insert the peg into the target hole, optionally with force feedback.
    \item The robot opens the gripper and retracts to a predefined configuration; the assembly outcome is inspected by the operator.
\end{enumerate}

\subsection{Metrics}

The $4$ metrics tracked in our study are described as follows. Other metrics like safety are beyond the scope of our study.

\begin{figure}%[!htb]
	\begin{subfigure}[t]{0.47\columnwidth}
		\centering
		\includegraphics[width=\textwidth]{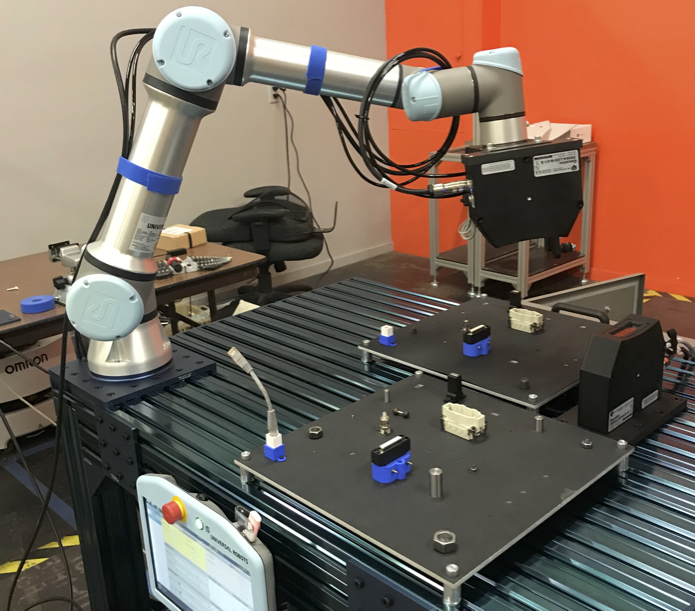}
		\caption{Horizontal placement}
	\end{subfigure}%
	\hfill
	\begin{subfigure}[t]{0.47\columnwidth}
		\centering
		\includegraphics[width=\textwidth] {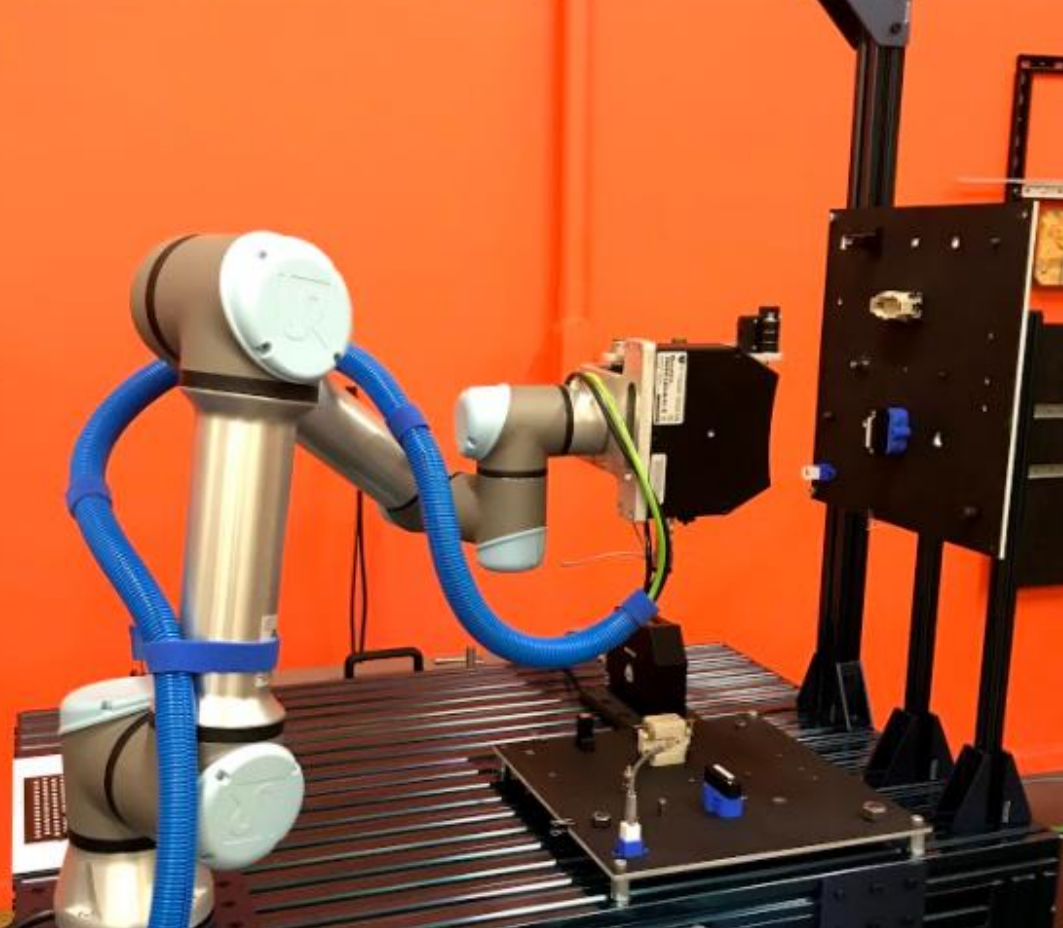}
		\caption{Vertical placement}
	\end{subfigure}
	\\[.5em]
	\begin{subfigure}[t]{0.47\columnwidth}
		\centering
		\includegraphics[width=\textwidth]{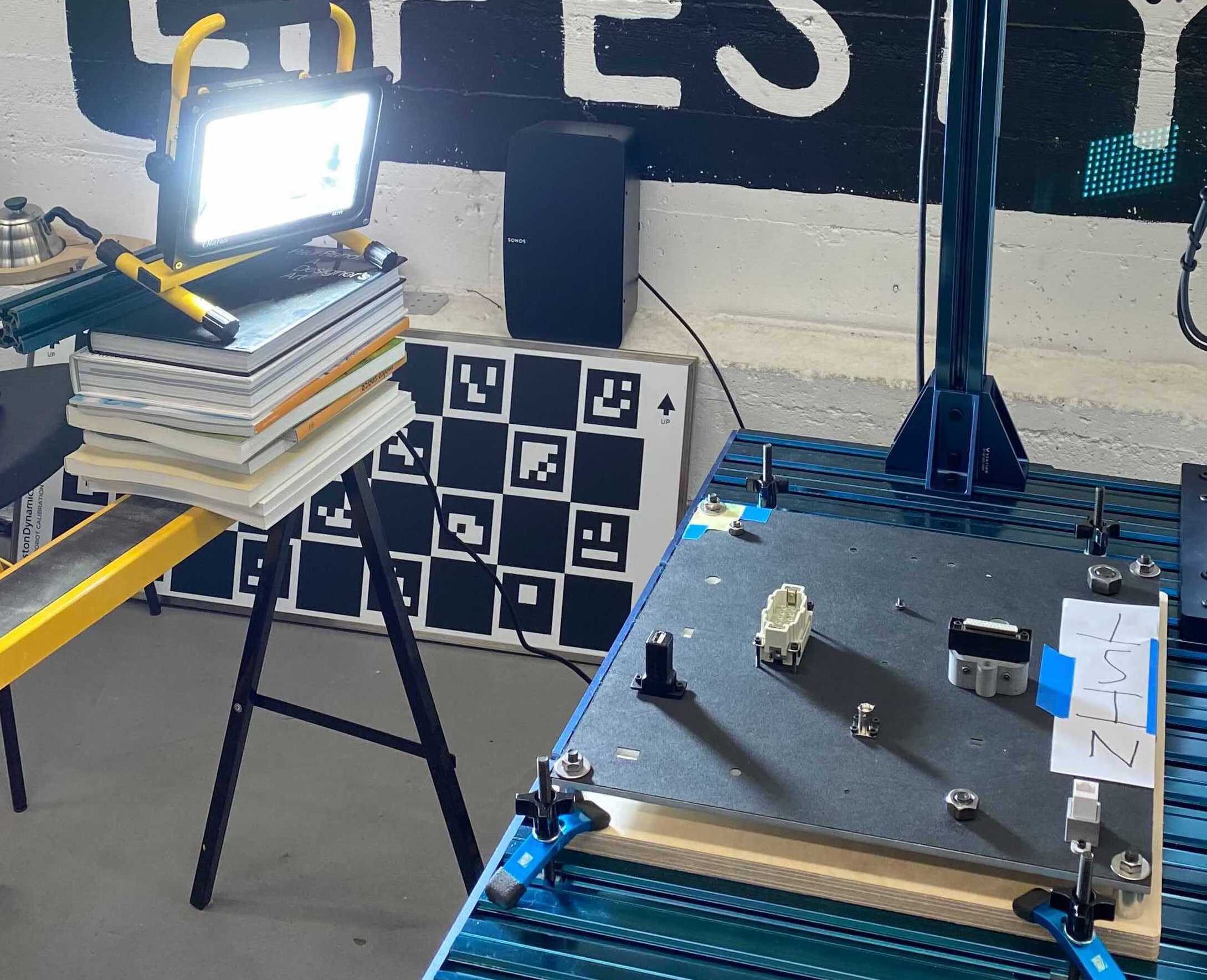}
		\caption{External lighting}
	\end{subfigure}%
	\hfill
	\begin{subfigure}[t]{0.47\columnwidth}
		\centering
		\includegraphics[width=\textwidth] {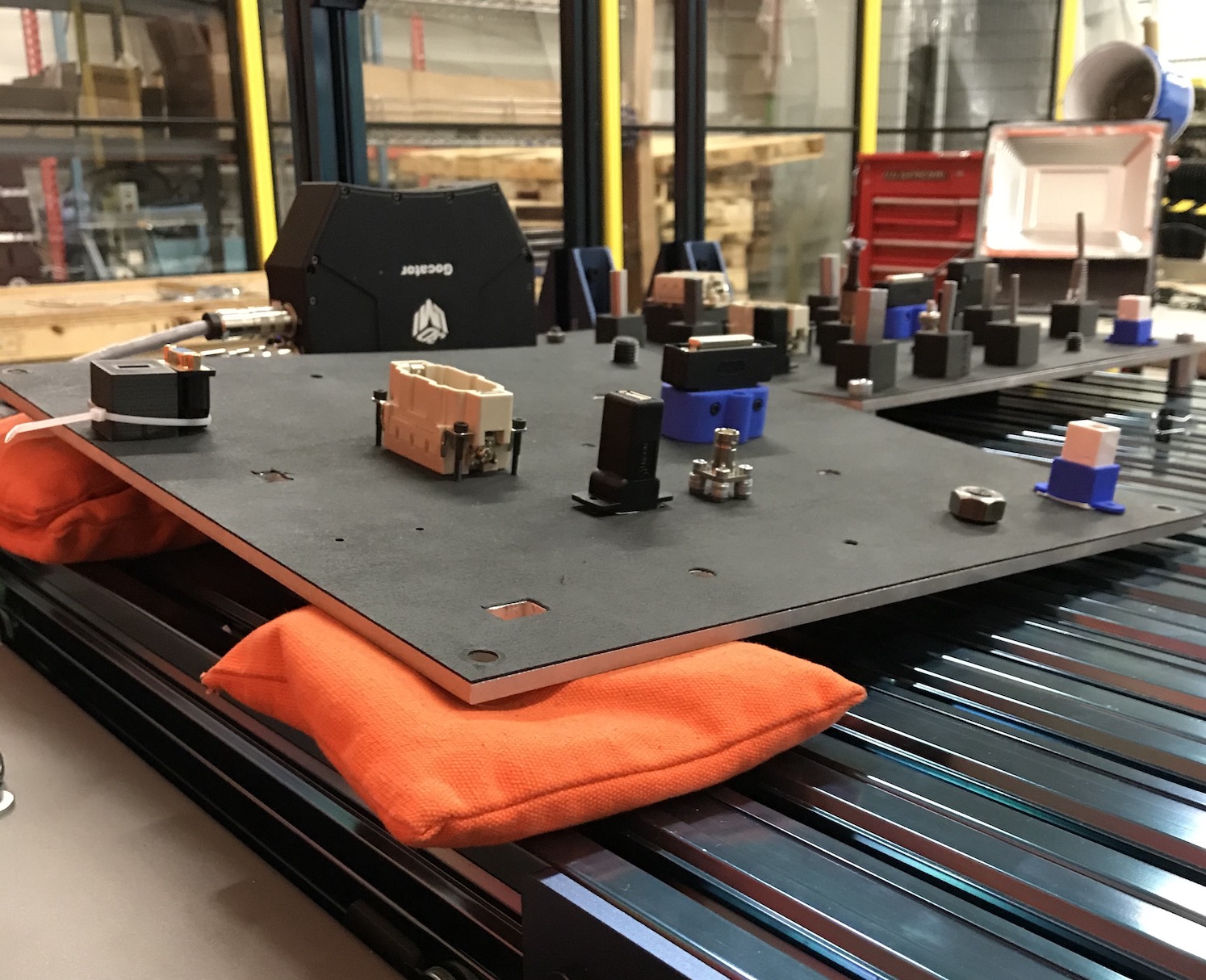}
		\caption{Movable placement}
	\end{subfigure}
	\caption{Task variations and setups. In (a, b, c), the assembly board is rigidly mounted on the worktable; while in (d), the board is pitched and resting on bean bags, movable during insertion.}
% \adam{Is it important to note that the board is padded, or just that it is pitched? T padded description may be confusing}
	\label{fig:experiment:task_variation}
	\vspace{-5mm}
\end{figure}

\textit{1) Reliability} measures the task completion capability of a solution under uncertainty. We characterize two kinds of uncertainty: task setup and task execution. Task setup variations are demonstrated in Fig.~\ref{fig:experiment:task_variation}, including whether the assembly board is placed vertically or horizontally, whether an external ambient light is turned on, and whether the assembly board is movable. Given a task setup, execution errors are unavoidable, e.g., the peg is grasped in a slightly different pose each time and the pose of the hole is imprecisely provided or estimated. Thus, we measure the end-to-end task completion rate at the task level and a \textit{tolerance area} at the module level, detailed in Section~\ref{section:result}.

\textit{2) Usability} is one determining factor on whether the solutions can be adopted by users without the domain expertise of solution developers. We measure usability by two numbers as follows. $T_{minimum}$ records the time an operator needs to adapt the OTS solution for a new task such that the robot achieves the first successful assembly. $T_{maximum}$ denotes $T_{minimum}$ plus the time the operator needs to fine tune the configurations such that the solution consistently accomplishes the assembly task, or no significant improvement can be obtained via additional tuning effort.

\textit{3) Cycle time} is recorded for reference but not optimized. A cycle starts when the robot begins to move towards the predefined grasp configuration, and ends after the robot makes an insertion attempt and retracts to a predefined configuration.

\textit{4) Cost} is a crucial factor affecting whether the systems are accessible to potential users. Considering the dynamic nature of pricing, we provide the price ranges of the relevant modules with our benchmarked solutions.

\section{Approaches}

There are a few crucial questions to investigate when designing the baseline study.
\begin{enumerate}
    \item Is there a significant difference from task to task, e.g., are some tasks much easier and error tolerant than others?
    \item How do force and vision based methods compare against each other?
    \item When will each solution break and in which scenarios does a selected solution perform reliably?
\end{enumerate}

\subsection{Insertion baselines}

As OTS vision software is commonly available for pose estimation via pattern matching at various levels of accuracy~\cite{hagelskjaer2019using}, most of our study focuses on the insertion stage assuming a known approximate, though not accurate, pose of the assembly board (equivalently, the target hole). Three solutions are implemented and tested on how they perform under this target pose uncertainty. At the end of the study, we selected a top-performing solution and integrated it with an OTS tool, to realize an end-to-end assembly solution.

\textit{1) Position control} (\texttt{Position}):
The robot moves to a predefined pose in the provided hole frame.

\textit{2) Hybrid motion/force control with spiral search}~\cite{abb_manual} (\texttt{Force}):
The robot moves to a predefined pose in the provided hole frame where it is close to establish contact. The robot moves straight along the assembly board's normal direction until the detected force is above a predefined threshold. The robot attempts to maintain a force $F_{normal}$ along the board's normal direction and 0 torque along all 3 axes, meanwhile, searching in a spiral pattern in the plane perpendicular to the board's normal direction. The procedure stops when the movement along the board's normal direction exceeds a threshold, namely, the pegs is considered being aligned with and inserted into the hole.

\textit{3) 3D vision guided position control} (\texttt{3DVision}): 
First, the robot moves to a predefined configuration for peg-in-gripper pose estimation using a static 3D camera. Second, based on the provided coarse estimate of the target hole pose, the robot moves such that the target hole is in the field of view of its wrist-mounted 3D camera and a fine estimate of the hole pose is obtained. Third, the robot moves to the predefined pose in the newly estimated hole frame. The pose estimation is done via template matching, i.e., a few 3D features are selected and a template is created for each part.

\subsection{System integration}

One drawback of OTS solutions we observed throughout this study is the lack of interoperatability, i.e., a hardware/software module is often found to be only compatible with a limited set of choices for other system components. One practical challenge encountered is, as an example, for a Universal Robot (UR) arm, there's no high precision hand-eye calibration tool available for the 3D camera we chose first and we ended up with a different camera. To ensure the generalizability of OTS solutions, different hardware components were used for the solutions studied.

For \texttt{Position}, a Kuka IIWA14 with a Robotiq Hand E gripper is used, though any other robot of similar control accuracy can be used\footnote{In our experiments, we empirically found the control accuracy is about 0.14mm in translation and 0.04 degrees in rotation.}.
% \adam{Can a similar statement be made for Force and 3DVision? Something to describe how applicable the results are. I.e., another robot system with XYZ capabilities could be used}.
For \texttt{Force}, an ABB IRB1200, a Robotiq 2F-85 gripper with customized fingers, an ATI Gamma ForceTorque sensor, and the ABB force package are used. For \texttt{3DVision}, a UR5e arm, a customized gripper, and two LMI 3506 Gocators are used (one is static and the other is wrist-mounted). For the final end-to-end solution, a Sensopart VISOR V20 is additionally integrated, which provides a template matching based pose estimation. A coarse cost estimate is provided in Table~\ref{table:systems}. 

For all solutions involving cameras, vendor-provided calibration routines are performed multiple times, to ensure a consistent hand-eye calibration. For both 3D cameras (LMI 3506 Gocators) and the 2D camera (Sensopart VISOR V20), a calibration error below $0.1$mm is obtained. This is significantly smaller than what is commonly seen in research projects, e.g., in the range of $0.5$mm to $5$mm. We hypothesize the calibration quality adds extra complexity in many research problems, aligned with the research directions proposed by \cite{lee2019camera, levine2016end}.

\begin{table}[h]
\caption{System components and cost estimate}
\label{table:systems}
\begin{center}
\begin{tabular}{@{}llr@{}} 
\toprule
Module & Method & Price Range\\
\midrule
6-axis force torque sensor &\texttt{Force}& \$5,000 - \$10,000 \\
Force control package &\texttt{Force}& \$5,000 - \$10,000 \\
Structured light camera & \texttt{3DVision} & \$10,000 - \$15,000 \\
Time-of-flight camera & End-to-End & \$5,000 - \$10,000 \\
\bottomrule
\end{tabular}
\end{center}
\vspace{-5mm}
\end{table}

\section{Experiments}
\label{section:result}

As we are focusing on the insertion stage, i.e., how each method is handling the residual error of the coarse target hole pose, in the majority of our experiments, we directly control the amount of pose error and investigate how reliable each method is. At the end of our study, a top-performing insertion method is integrated into an end-to-end system, demonstrating the capability of solving the complete assembly tasks.

\subsection{Intrinsic task difficulty}
\label{section:task_difficulty}
The 12 tasks considered are diverse in shape, dimension, and clearance, hence, it is crucial to establish an understanding how difficult each task is. In this section, we first analyze how the baseline \texttt{Position} performs. We follow the procedure outlined in Section~\ref{section: tasks}, except in Step 2), the peg is inserted
% \adam{maybe clarify that this is the starting position of the peg} 
into its corresponding hole (instead of the nest) as the starting state for grasping, to minimize the peg-in-gripper pose uncertainty. The nominal target hole pose is defined such that the peg and the hole are perfectly aligned for accomplishing the task. Then we inject a 3D perturbation error: along the two axes in translation in parallel with the assembly board plane ($x, y$), and in rotation around the assembly board's normal direction ($rz$). It is noteworthy that depending on the clearance level of each peg-hole pair, a small rotation error out of the board plane might exist ($rx$ and $ry$). For instance, the rectangle pegs can be slightly tilted when inserted as their lengths are much larger than the the board thickness.

We aim to estimate a \textit{tolerance area} for each task approximating the intrinsic task difficulty, i.e., the 2D area within which insertion succeeds, robust to the injected error. One practical challenge that peg-in-hole benchmarking encounters is that wear-and-tear quickly changes the difficulty of particular tasks, i.e., insertion becomes increasingly easier with more attempts, which cannot be neglected for the open loop position control method. Avoiding the otherwise laborious procedure of switching tens of assembly boards, we used two boards with one for coarse benchmarking and the other for validation.\footnote{Using aluminum boards instead of plexiglass boards as in the original NIST Assembly Task Board \#1 also helped reducing wear-and-tear effects.}

Specifically, on the first board, we start the insertion attempts using the nominal pose, repeating the aforementioned procedure for 10 times with $rz$ rotation error smaller than 2 degrees. Then we increase the translational error by 0.5mm along $x$ and/or $y$ axis following an outwards pattern on a 2D grid, and repeat 10 attempts at each grid point. This process continues until consecutive failures are observed. Grid points with all successes are labeled as \textit{succeeded} and the rest grid points with nonzero successes are labeled as \textit{possible}. Then we use the second board to validate this tolerance map to mitigate the effect of wear-and-tear. For validation, we conduct 2 attempts on each of the non-failure points and update the tolerance map: \textit{succeeded} points are updated to \textit{possible} if one failed and \textit{failed} if both failed; \textit{possible} points are updated to \textit{failed} if both failed.
% \dirk{wondering if we should really add one (or all) of the grid plots, both for further detail and more clarity here}

\begin{figure}%[!htb]
	\centering
	\includegraphics[width=\columnwidth]{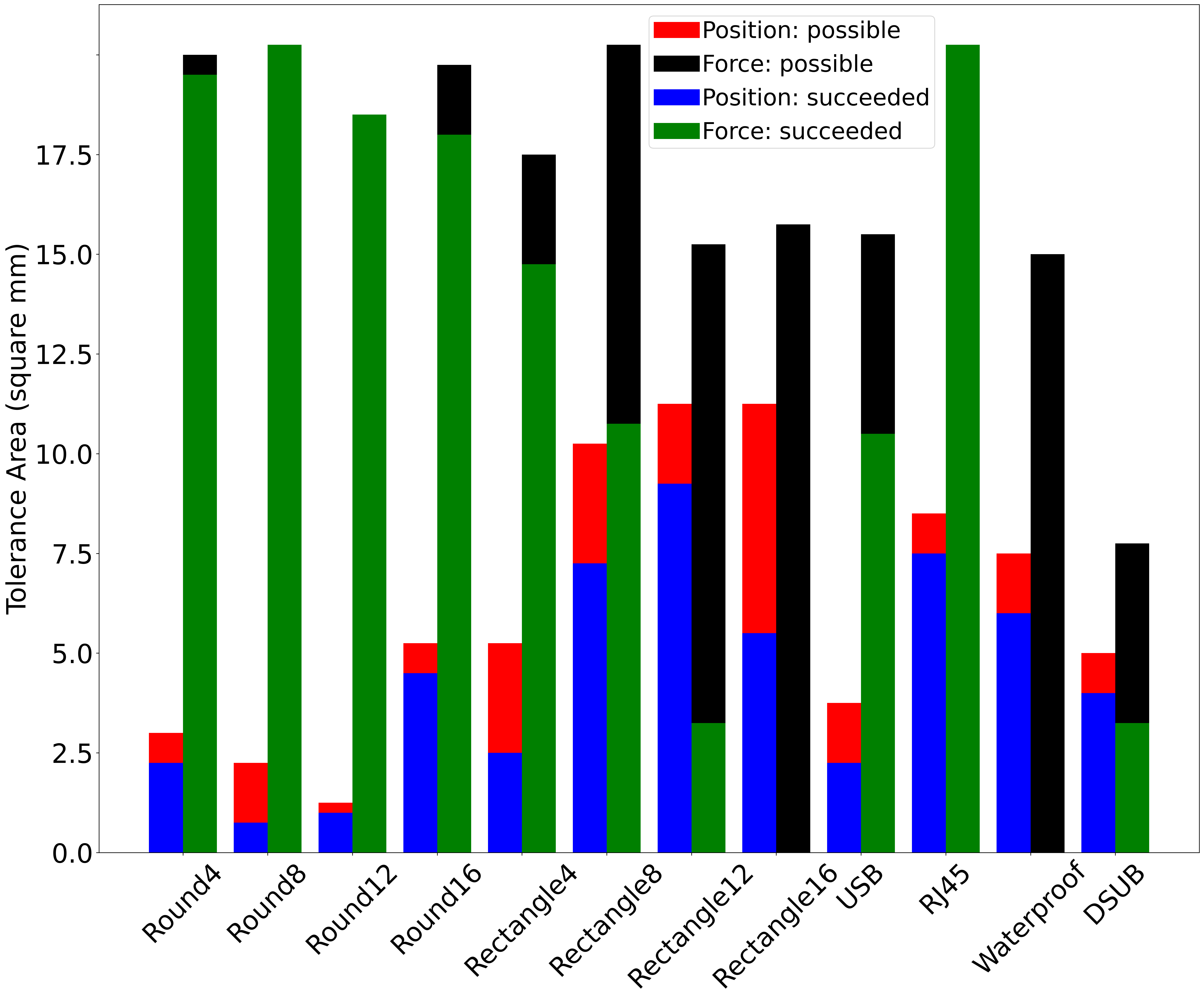}
	\caption{Tolerance areas of 12 insertion tasks.}
	\label{fig:experiment:tolerance_map}
% 	\dirk{increase font size, use colors that work on b/w, upper case Power/Force}
\end{figure}

As summarized in Fig.~\ref{fig:experiment:tolerance_map}, different tolerance levels are observed across the 12 tasks. Factoring out the dimension impact, compared with the round pegs, rectangle pegs have high tolerance thanks to their chamfered edges. In contrast to rectangle pegs, the tolerance areas of round pegs are not correlated with their dimensions, but determined by the manufacturing precision. Among the connectors, USB has the lowest tolerance because of its small dimension and sharp edges. RJ45 has the largest tolerance because of the large clearance between the connector end and the socket opening. Waterproof also has a large tolerance area as the socket has round edges and is subject to minor deformation.

\subsection{Sensor-based methods}
\label{section:sensor_based}

\subsubsection{Reliability}
We first compare how \texttt{Force} and \texttt{3DVision} perform in terms of reliability. As seen from the results in Section~\ref{section:task_difficulty}, position control without sensor feedback fails when the target hole error is above some task-dependent threshold. For \texttt{Force}, we follow the same protocol but cap the injection error range at maximum 2mm along $x$ and $y$ axis\footnote{We assume this level of precision can be achieved by an upstream vision module or a fixture system at low cost.}. The result is demonstrated in Fig.~\ref{fig:experiment:tolerance_map} in comparison with \texttt{Position}.

It can be observed that \texttt{Force} increases the tolerance area combining \textit{succeeded} and \textit{possible} across all peg types. The performance in round pegs, in particular, exceeds \texttt{Position} significantly; this is because the peg gets caught and aligned with the hole almost surely during spiral search due to the simplistic geometry. However, the tolerance area of \textit{succeeded} decreases in some of the peg types, especially for the connectors such as the waterproof and the DSUB, i.e., \texttt{Force} enables the robot to handle larger upstream errors but decreases the reliability. Two major error patterns causing this reliability degradation are detailed below.

\begin{figure}%[!htb]
	\centering
	\includegraphics[width=\columnwidth]{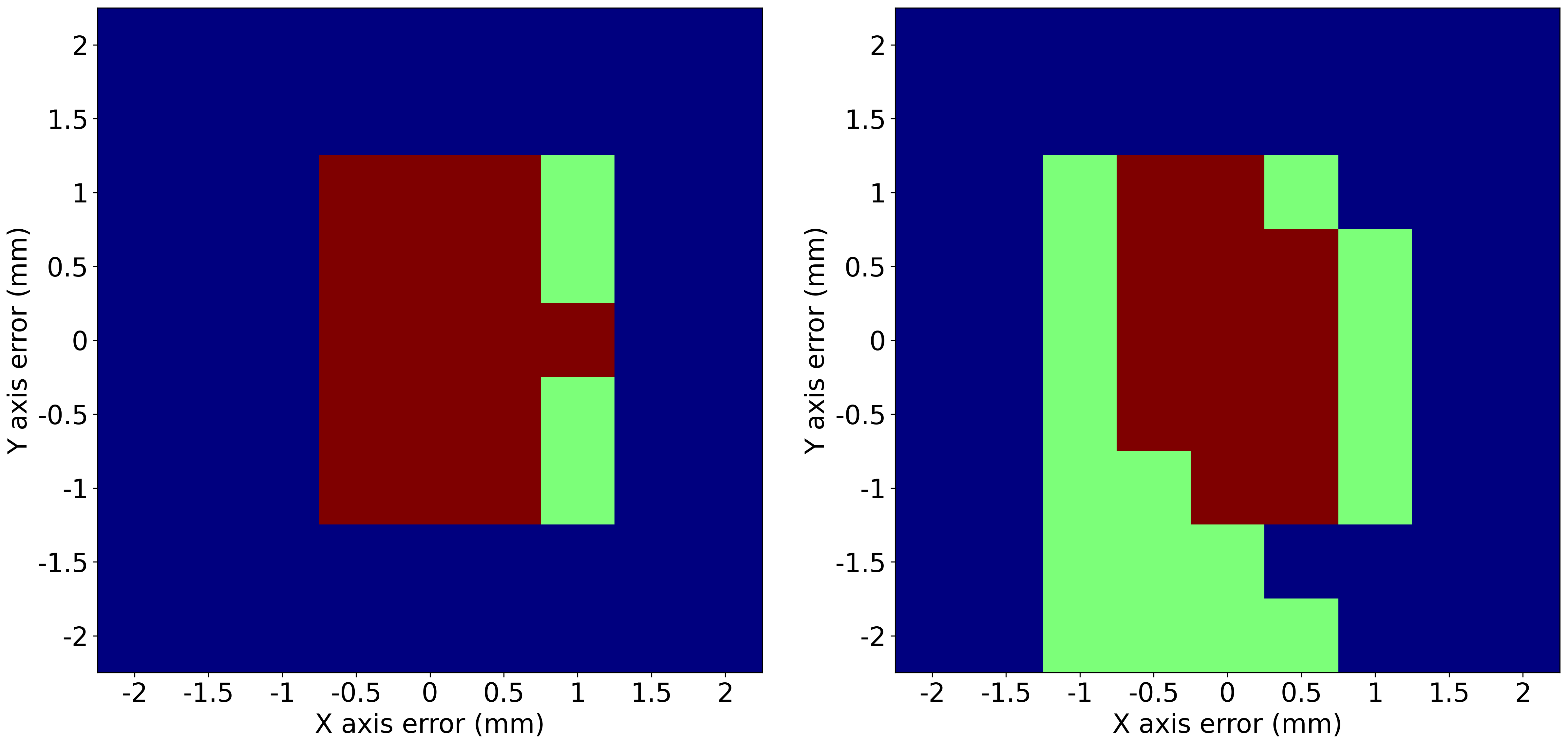}
	\caption{Tolerance map of DSUB insertion. Red denotes \textit{succeeded}, green denotes \textit{possible}, and blue denotes \textit{failed}. Left: \texttt{Position}, right: \texttt{Force}.}
	\label{fig:experiment:dsub_map}
	\vspace{-2mm}
\end{figure}

First, spiral search does not tackle the rotational error. Therefore, the pegs with sharp edges and corners can easily get jammed and there is no anomaly detection or error recovery mechanisms. This type of failure cases is often observed in rectangle peg and connector insertion tasks. Second, as the robot attempts to maintain zero torque during search, it relies on some non-flat features to lock the peg such as the hole edge or socket boundary extrusion. While this assumption holds in most peg-in-hole tasks, there are ``false positive'' non-flat features such as tiny bumps due to manufacturing precision, material wear-and-tear, or simply, the parts have non-flat features by design. One typical example is illustrated by the DSUB tolerance map in Fig.~\ref{fig:experiment:dsub_map}, where the connector pins are often latched even the connector and socket are not aligned.

% The spiral search pattern is governed by the spiral velocity and radius. Consequently, if the settings aren't configured properly for a peg-hole pair, some initial misalignment configurations cannot be recovered from the search. This happens during the round pegs experiments where failures were observed at a few locations on the 2D map. This issue can be addressed by more parameter tuning or ideally, self-optimization. 

The same experimental protocol as described in Section~\ref{section: benchmark} is followed for \texttt{3DVision}, except that we introduce workspace variations for Step 1) to further test the vision module robustness, including vertical assembly board placement and extra lighting (Fig.~\ref{fig:experiment:task_variation} b and c). 150 trials are attempted for each insertion task and a $\mathbf{100\%}$ task completion rate is achieved except 4 failures observed on RJ45. The tolerance area is practically determined by its field of view (22mm in $x$, 33mm in $y$, and 18mm in $z$ for the 3D camera we used).

\begin{figure}%[!htb]
	\centering
	\includegraphics[width=0.48\columnwidth]{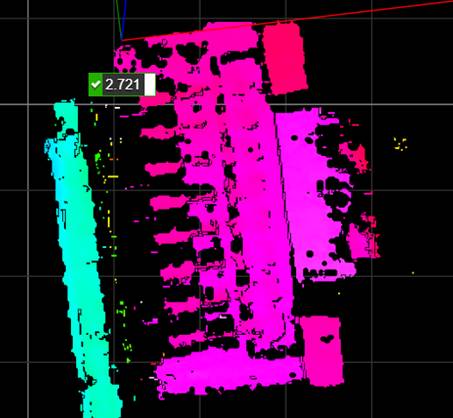}
	\hfill
	\includegraphics[width=0.48\columnwidth]{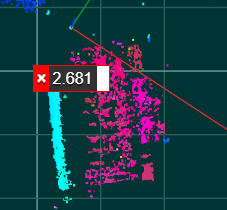}
	\caption{Example point cloud with the detection result of the RJ45 connector. The detected pose is plotted as xyz axes, colored with RGB. Left: success; right: failure caused by missing point cloud measurements. The green check and red cross in the figures are validation features provided in the vision tool, not used in our study.}
	\label{fig:experiment:rj45_3d}
	\vspace{-3mm}
\end{figure}

We then take a detailed look at the failure scenarios. Illustrated in Fig.~\ref{fig:experiment:rj45_3d} are two example point cloud measurements along with the detection results of the RJ45 connector. As the pattern matching algorithm relies on 3D geometric features such as 3D edges and circles, it is subject to measurement noise, particularly on translucent surfaces like RJ45 connectors. In the right panel of Fig.~\ref{fig:experiment:rj45_3d}, as the points in the connector's top-right region are missing, the 3D edge is wrongly fit, giving a false x-axis (shown in the red). While in the left panel, the point cloud is mostly complete, and correct 3D edges are found thus generating an accurate pose.

\subsubsection{Usability}
Table~\ref{table:usability} summarizes the effort required to get familiar with the OTS tool and further, to customize settings to achieve satisfying performance. As seen in the table, \texttt{Position} does not require much effort adapting to a new task besides setting a number of waypoints. There is no further effort needed for fine tuning.

\texttt{Force} requires the operator to set the target frame, along with some parameters including the spiral search velocity, the search radius, the force threshold, the search stopping criteria, and the stiffness~\cite{abb_manual}. In our study, we found that it is relatively easy to achieve the first task success. However, fine tuning the parameters to achieve satisfying reliability remains challenging, which has to be done via trial-and-error. Interacting with workpieces causes wear-and-tear and, in turn, changes the task itself. Sample efficient methods for parameter optimization, leveraging simulation, and non-invasive contact-rich manipulation learning can be promising directions to explore to address such issues.

\texttt{3DVision} requires a decent amount of effort to configure the desired peg-hole relative poses in the camera frame, adjust camera configuration such as the exposure time, and pose estimation algorithm parameters. Getting the first task success takes on average 2 hours in our study but more trial-and-error is needed to fine tune the pose estimation setting, e.g., filter kernel size for outlier rejection and finding robust 3D features. For the assembly board studied, as there is not much background clutter, we found it's tractable for users without domain expertise to achieve satisfying reliability using the vendor-provided software and procedure.

\begin{table}[h]
\centering
\caption{Usability comparison}
\label{table:usability}
\begin{tabular}{@{}lSS@{}} 
\toprule
& \multicolumn{2}{c}{Time in hours (h)} \\ 
\cmidrule(r){2-3}
Method & $T_{minimum}$ & $T_{maximum}$ \\
\midrule
\texttt{Position} &  <0.5 & <0.5 \\
\texttt{Force} & \sim 0.5 & >5 \\
\texttt{3DVision} & \sim 2 & \sim 5 \\
\bottomrule
\end{tabular}
\vspace{-3mm}
\end{table}

\subsubsection{Cycle time}

In our benchmarking study, cycle time is not optimized to compete with the human baseline provided in~\cite{falco2018performance}. We provide the average and standard deviation of cycle times across all 12 tasks for reference, as shown in Table~\ref{table:cycle_time}. The cycle time is computed excluding the grasping stage, i.e., the time starts after the peg is grasped in the gripper. Many steps can be improved such as optimizing trajectories for all methods, fine tuning the searching velocity for \texttt{Force}, using powerful computers instead of onboard processing and processing images during robot movement for \texttt{3DVision}.

\begin{table}[h]
\centering
\caption{Cycle time comparison on insertion tasks}
\label{table:cycle_time}
\begin{tabular}{@{}lSSS@{}} 
\toprule
& \multicolumn{2}{c}{Time in seconds (s)} \\ 
\cmidrule(r){2-3}
Method & \multicolumn{1}{c}{Mean cycle time} & \multicolumn{1}{c}{Standard deviation} \\
\midrule
\texttt{Position} & 3.3 & 0.1\\
\texttt{Force} & 10.7 & 4.2\\
\texttt{3DVision} & 18.7 & 2.4\\
\bottomrule
\end{tabular}
\vspace{-5mm}
\end{table}

\subsection{End-to-end solution}

From the analysis on all the insertion methods in Section~\ref{section:sensor_based}, it can be concluded that \texttt{3DVision} achieves the highest reliability. Thus we integrate this method with a coarse target pose estimation module and arrive at our end-to-end OTS solution.

\begin{figure}%[!htb]
	\centering
	\includegraphics[width=0.24\columnwidth]{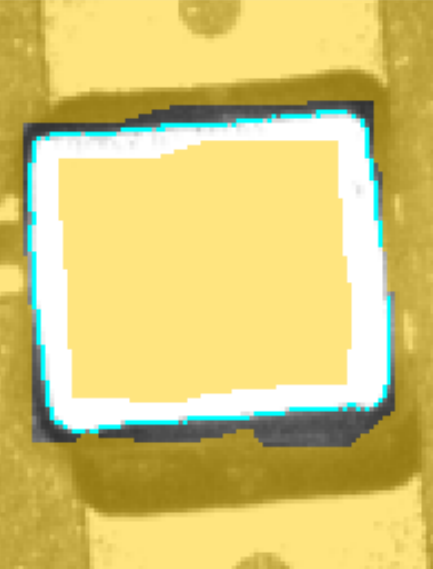}
	\hfill
	\includegraphics[width=0.205\columnwidth]{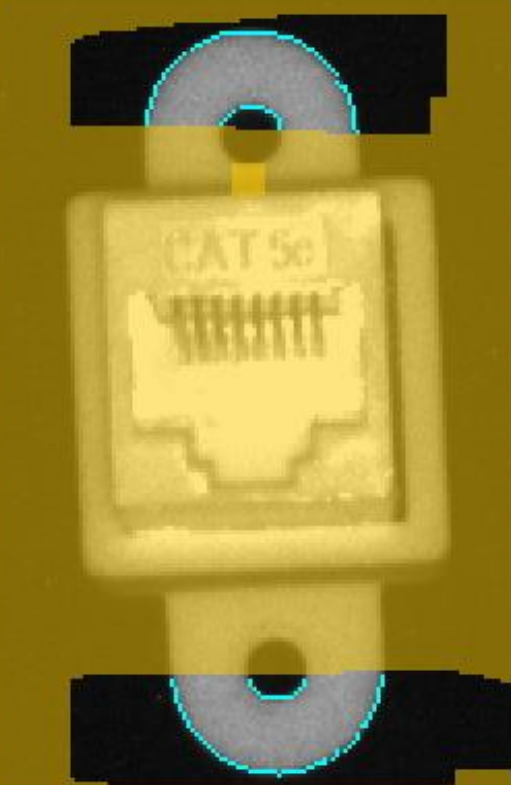}
	\hfill
	\includegraphics[width=0.208\columnwidth]{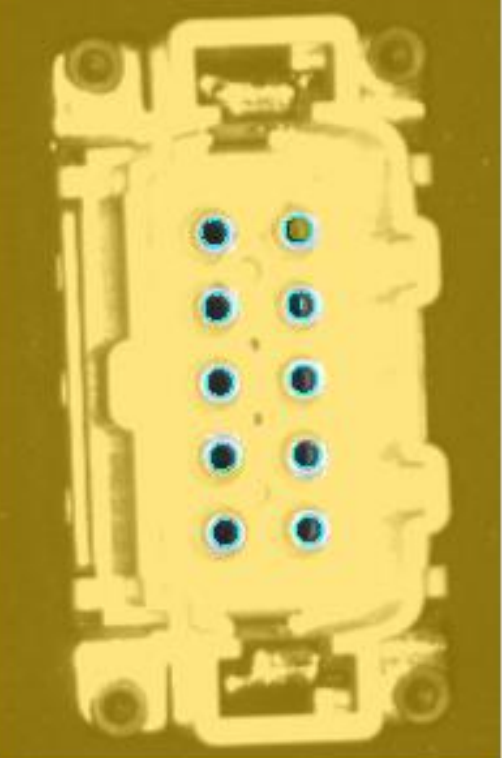}
	\hfill
	\includegraphics[width=0.24\columnwidth]{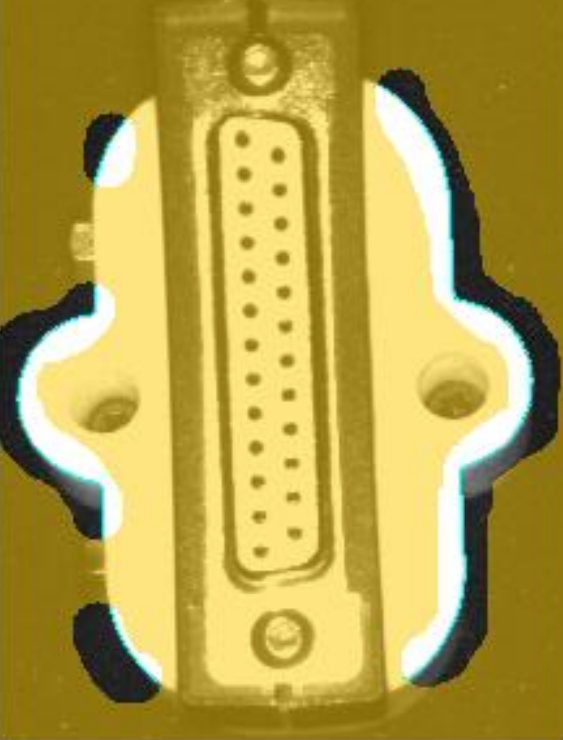}
	\caption{Example 2D socket features (in cyan) used in coarse target pose estimation. Left to right: USB, RJ45, Waterproof, DSUB.}
	\label{fig:experiment:2d_features}
\end{figure}

To estimate the coarse target hole pose, we use the Sensopart VISOR V20 camera along with its image processing package. The camera is equipped with a monocular sensor and a time-of-flight module. 2D pattern matching is used for planar feature localization and the depth data is used to derive the 3D information (the hole center coordinate and normal direction). Example 2D features used in our experiments are demonstrated in Fig.~\ref{fig:experiment:2d_features}, where the color cyan denotes the chosen features such as curves for DSUB and circles for Waterproof.

A separate accuracy test is performed first to verify this vision module meets the requirement from downstream insertion tasks. Given the fact that the high-precision 3D camera tested in Section~\ref{section:sensor_based} achieves sub-millimeter accuracy, we use its detection result as the ground truth and evaluate the coarse pose estimation performance. The observed maximum translation error across all tasks is around 6mm, seen on the USB socket as most surfaces of the socket were black rendering low contrastness. The maximum orientation error is about 9 degrees, seen on the Waterproof socket because the circular features do not compose a perfect plane and are subject to noise. This error range confirmed the validity of our choice of \texttt{3DVision}, which has a larger tolerance area as mentioned in Section~\ref{section:sensor_based}.

We finally benchmarked the end-to-end solution on all connectors (i.e., USB, RJ45, Waterproof, and DSUB) using the procedure described in Section~\ref{section: benchmark}, with 50 trials each. The assembly board is placed horizontally and movable with bean bags beneath altering its pitch, as shown in Fig.~\ref{fig:experiment:task_variation}d; cables are attached with the connectors during experiments to be closer to realistic settings.
% \adam{Is there room for an image of this? Or is this shown in Fig. 2?}.
The performance is summarized in Table~\ref{table:end_to_end}. The RJ45 insertion is not robust due to the incomplete point cloud issue discussed in Section~\ref{section:sensor_based}. One thing to note is that attaching cables to connectors does cause failures in experiments. Specifically, there were cases when a cable segment was grasped together with the connector by the closed fingers, leading to a wrongly estimated connector-in-gripper pose, thus failing the later insertion. This suggests the need for 1) a more robust vision algorithm that handles unseen corner cases, and 2) scene awareness which reasons semantically along with geometric understanding. Another area not covered in this study is failure recovery, i.e., a second attempt can be performed if anomaly is detected.

\begin{table}[h]
\centering
\caption{End-to-end performance of connector insertion}
\label{table:end_to_end}
\begin{tabular}{@{}lSS@{}} 
\toprule
Connector Type & \multicolumn{1}{c}{Task completion rate (\%)} & \multicolumn{1}{c}{Mean cycle time (s)} \\
\midrule
USB & 100.0 & 29.8 \\
RJ45 & 96.0 & 33.6 \\
Waterproof & 98.0 & 30.1 \\
DSUB & 100.0 & 28.0\\
\bottomrule
\end{tabular}
\end{table}

It is worthwhile to remark that the vision tools are subject to failures as they rely on low level vision cues such as edges and circles, and they are designed to work only in a confined workspace without background clutter. If the workspace becomes less controlled or structured, false positives emerge caused by distractors such as other parts and cables, and false negatives also occur from factors including lighting change and part reflection.

\section{Conclusions}

In this paper, we designed a customized assembly benchmark leveraging the recently proposed NIST Assembly Task Boards and provided an analysis on the task intrinsic difficulty. Multiple off-the-shelf (OTS) sensor-based methods were integrated and studied, including hybrid motion/force control and vision-based pose estimation. Experiments demonstrated that the OTS solutions are capable of solving the assembly tasks with high success rate, except in known problematic scenarios such as cable interference and transparent parts. It is noteworthy that the tasks studied do not require extreme precision in the range of micrometers, thus much less error tolerant due to jamming. Also, grasping objects from random poses in cluttered bins is not addressed in our study, which is deferred to future work.
% We also did not address grasping from random start poses, as it in bin-picking.

A few research directions are suggested: 1) improving usability, reducing domain expertise required from users and simplifying the procedure adapting to new tasks; 2) increasing applicability, enabling the assembly system to work with broader categories of workpieces and reducing corner cases; 3) improving interoperability, designing software compatible with a wider range of hardware; 4) developing scene awareness and failure recovery mechanisms, empowering the robot to have realtime awareness of its own and surrounding geometric and semantic state; 5) lowering the cost, allowing the use of cheaper hardware.

\bibliographystyle{IEEEtran}
\bibliography{references}

\end{document}